%% file: Paper-2687.tex
\documentclass[runningheads]{llncs}
\usepackage[T1]{fontenc}
\usepackage{graphicx,verbatim}
\usepackage{wrapfig}
\usepackage{graphicx}   

\input{preamble}

\begin{document}
\title{\ours{}: Bin‑wise Certification of Confidence Calibration in Medical Image Classification}
\titlerunning{\ours{}: Bin‑wise Certification of Confidence Calibration}

%

\author{Leo Fillioux\textsuperscript{1*}, Stergios Christodoulidis\textsuperscript{1}, Paul-Henry Cournède\textsuperscript{1}, Maria Vakalopoulou\textsuperscript{1} and Jose Dolz\textsuperscript{2}}  
\authorrunning{L. Fillioux et al.}
\institute{\textsuperscript{1}Université Paris-Saclay, CentraleSupélec, Gustave Roussy, INSERM, CDSU, IHU PRISM, Gif-sur-Yvette\\
\textsuperscript{2}LIVIA, ILLS, ETS Montréal\\
    Corresponding author: \email{leo.fillioux@centralesupelec.fr}}
  
\maketitle              
%

\begin{abstract}
Deep neural networks remain vulnerable to adversarial perturbations, which can distort not only predictions but also confidence scores, undermining uncertainty calibration. While existing certification methods focus on preserving the predicted category, providing guarantees on how calibration behaves under adversarial attacks remains overlooked. In this work, we introduce \ours{}, a simple and efficient post‑hoc strategy that certifies bin‑wise confidence calibration for any pretrained differentiable classifier. Our approach combines empirical calibration estimates, statistical concentration bounds, and local Lipschitz estimates of the confidence function to derive data‑dependent upper bounds on worst‑case miscalibration within an $\ell_2$-ball of radius R. We evaluate \ours{} across 11 medical image classification tasks and multiple adversarial perturbations, demonstrating substantially higher certified coverage than baseline strategies while maintaining competitive tightness. Our code is available at \href{https://github.com/leofillioux/calcert}{github.com/leofillioux/calcert}.
\blfootnote{\textsuperscript{*} Work done during an internship at ILLS.}
\end{abstract}

\section{Introduction}
\label{sec:introduction}
Adversarial attacks can drastically alter the predictions of modern neural networks, even when perturbations are visually imperceptible \cite{szegedy2013intriguing}. Consequently, models that appear robust under current evaluation protocols may remain vulnerable to these attacks. Moreover, label attacks \cite{goodfellow2014explaining,kurakin2018adversarial,madry2018towards,shafahi2018poison}, the most prominent form of adversarial perturbations, affect not only prediction correctness, but also the confidence scores, thereby degrading overall uncertainty calibration. This is particularly concerning in safety-critical domains, such as medical image analysis \cite{bortsova2021adversarial,ma2021understanding}, where confidence scores often play a central role in downstream decision-making, and where failures can have severe real-world consequences. 

Given these risks, there is a growing need for formal guarantees on model behavior under perturbations. While adversarial training and empirical defenses can improve robustness, they provide no provable guarantees and are frequently broken by stronger attacks. In contrast, certified robustness \cite{carlinicertified,cohen2019certified,singh2019boosting} has emerged as a promising direction, providing mathematical guarantees that a model's output remains stable for all inputs within a prescribed $\ell_p$-norm ball of radius $R$. Nevertheless, computing exact adversarial robustness is coNP-hard \cite{katz2017reluplex}, and attack-based evaluations provide only heuristic upper bounds on vulnerability \cite{carlini2017towards,tramer2020adaptive}. This has motivated a rich and growing line of work on certification methods, such as Randomized Smoothing~\cite{carlinicertified,cohen2019certified}, which provide provable guarantees on robust accuracy. Yet, despite this progress, existing certification frameworks focus almost exclusively on providing guarantees that the model predicted class remains unchanged \cite{hussein2024promptsmooth,laousy2023certification}
given a prescribed perturbation radius, but they do not address how confidence scores behave. This is a critical omission, as robustness of accuracy does not directly imply robustness of calibration, and a model may remain correct while becoming arbitrarily over‑ or under‑confident.

To our knowledge, certifying uncertainty calibration in the medical domain remains unexplored. Existing method provides no guarantees on how calibration may deteriorate under adversarial perturbations, or how bin-wise miscalibration behaves. This gap is critical in healthcare, where confidence (rather than accuracy alone) drives downstream actions, motivating the question: 
\textit{Can we certify the uncertainty calibration of a pretrained model under $\ell_2$-bounded perturbations without retraining or attack-specific fine-tuning?}

Thus, we can summarize our main contributions as:
\begin{itemize}
\item We take a first step toward certifying uncertainty calibration in medical image classification under adversarial perturbations. To this end, we derive a tractable, data-dependent upper bound on the worst-case change in confidence within an $\ell_2$-ball by estimating local Lipschitz constants of the model confidence function. 
This provides a principled way to control how far confidence values can drift under input perturbations.

\item Building on this, we obtain certified upper bounds on \emph{bin-wise} miscalibration by combining empirical calibration estimates, statistical concentration inequalities, and our Lipschitz-based control of confidence variation. The resulting certificate is post-hoc, requires no adversarial training and is compatible with any differentiable model. This makes our approach simple to implement and applicable to a wide range of pretrained classifiers.
\item Extensive experiments across 11 classification tasks spanning multiple medical imaging modalities, five diverse perturbations, including adversarial attacks, and four backbones (leading to a total of 220 different settings) demonstrate that \ours{} yields usable, non-vacuous bounds for calibration certification.
\end{itemize}

\section{Related Work}
\label{sec:related_work}

\noindent \textbf{Certified robustness for deep networks} aims at guaranteeing that model predictions remain stable under bounded adversarial perturbations. Empirical defenses such as adversarial training \cite{athalye2018obfuscated,engstrom2019exploring,ghiasibreaking} can improve robustness in practice but offer no formal guarantees. Certified defenses instead provide provable robustness under specific threat models, most commonly via \textit{randomized smoothing} (RS) \cite{carlinicertified,cohen2019certified,mohapatra2020higher,rosenfeld2020certified,salman2020denoised} or Lipschitz-controlled architectures \cite{cisse2017parseval,fazlyab2023certified,huang2021training,zhang2022rethinking}. RS \cite{rosenfeld2020certified} constructs a Gaussian‑smoothed classifier and yields probabilistic guarantees on robust accuracy. However, it certifies only label invariance, i.e., no guarantees on confidence or calibration, and requires retraining or noise-specific tuning, making it unsuitable for our goal of post-hoc calibration analysis. \textit{Lipschitz-based certification methods} \cite{cisse2017parseval,fazlyab2023certified,huang2021training,zhang2022rethinking} enforce global or local Lipschitz constraints during training, limiting output variation under bounded perturbations. However, they require architectural or training modifications and are therefore not compatible with our post-hoc, model-agnostic setting.

\noindent \textbf{Robustness certification in medical imaging.} Providing accurate and reliable models is essential in medical imaging, especially as AI systems increasingly receive regulatory approval (e.g., FDA, CE) and are deployed in clinical practice. Thus, recent work has begun exploring certified robustness in this domain. \cite{laousy2023certification} presented a method to verify robustness in the context of medical imaging  segmentation based on RS and diffusion models. More recently, \cite{hussein2024promptsmooth} used diffusion- and prompt-based smoothing to improve robustness to noise and distribution shift in medical vision systems. These advances highlight that certified robustness is becoming a critical component of trustworthy medical imaging systems and represents an important step toward safe and reliable clinical deployment. 

Unlike prior work, we focus on \emph{certifying confidence calibration} under adversarial perturbations, a problem still unexplored in medical imaging, and take a first step toward extending certification beyond accuracy to confidence reliability.

\section{Methodology}
\label{sec:methodology}

\subsection{Problem setup}
Let $f_\theta : \mathbb{R}^{d} \to [0,1]^C$ be $C$-class classifier with predicted probabilities $p(x) = f_\theta(x)$, where the  predicted label is defined as 
$\hat{y}(x) = \arg\max_{c} p_c(x)$, and whose associated confidence is $p_{\hat{y}(x)}(x)$.  
A classifier is said to be well calibrated if its predicted confidence matches the true likelihood of correctness. Formally, for a prediction with confidence $p_{\hat{y}(x)}(x)$, the model is calibrated when $\mathbb{P}\bigl(y = \hat{y}(x) \,\big|\, p_{\hat{y}(x)}(x) = t\bigr) = t$ for all $t \in [0,1].$ In practice, this conditional relationship is approximated by using a binning scheme: the confidence interval $[0,1]$ is partitioned into $K$ bins $B_k = [a_k, b_k],$ and for each bin we compute empirical accuracy $\hat{A_k}$, mean confidence $\hat{P_k}$, and miscalibration $\widehat{\mathrm{Cal}}_k = |\hat{A}_k - \hat{P}_k|,$ based on a given calibration dataset $\mathcal{D}_{cal} = \{(x^{(n)}, y^{(n)})\}_{n=1}^N.$ Our goal is to bound how much bin-wise calibration may change under additive $\ell_2$-bounded perturbations. 

\subsection{Proposed confidence certification}

A central difficulty of this problem is that, under perturbations $\|\delta\|_2 \le R$, both the predicted confidence and the predicted label may change in ways that are hard to characterize directly. To obtain certified guarantees, we therefore require an upper bound on how much the model confidence can vary within the perturbation ball. Since evaluating the classifier on all possible perturbed inputs is infeasible, we approximate this worst-case change using a local Lipschitz bound on the confidence function. This provides a tractable way to control the effect of perturbations and forms the basis of our certified calibration analysis.

\noindent \textbf{Local Lipschitz constants for confidence.} For a fixed input $x$, let us define the scalar confidence
function $g(x) = p_{\hat{y}(x)}(x).$ Then, we approximate its local Lipschitz constant by the gradient norm $L^{\text{local}}(x) \approx \|\nabla_x g(x)\|_2$, computed via automatic differentiation by treating $g(x)$ as a scalar objective\footnote{This provides a tractable,
first-order approximation of the local Lipschitz constant needed for the
certified calibration bound.}. Under this approximation, any perturbation $\delta$ with 
$\|\delta\|_2 \le R$ satisfies
\begin{equation}
\label{eq:lip_bound}
  |g(x+\delta) - g(x)|
  \le L^{\text{local}}(x)\,R,
\end{equation}
which provides a pointwise, first-order certificate on how much confidence can change within the perturbation ball.

\noindent \textbf{Bin-wise Lipschitz constants.} We adopt the standard binning strategy for calibration, partitioning the confidence interval $[0,1]$ into $K$ bins $B_k = [a_k, b_k].$ 
To obtain certified guarantees, we must ensure that samples assigned to a bin remain in that bin under all admissible perturbations. For each bin 
$B_k$, we define a \emph{robust sub-bin} $\widetilde{B}_k = [a_k + \gamma_k,\; b_k - \gamma_k],$ where $\gamma_k > 0$ is chosen so that any sample whose confidence lies in $\widetilde{B}_k$ cannot cross a bin boundary under perturbations of size $R$. Using the Lipschitz bound in \eqref{eq:lip_bound}, a sufficient condition is:
\begin{equation}
\label{eq:gamma}
  \gamma_k \ge L_kR \ge
  \max_{n \in \widetilde{\mathcal{I}}_k}
  L^{\text{local}}(x^{(n)})\,R,      
\end{equation}
where $\widetilde{\mathcal{I}}_k = \{n : p_{\hat{y}(x^{(n)})}(x^{(n)}) \in \widetilde{B}_k\},$ and the bin-wise Lipschitz constant is $L_k = \max_{n \in \widetilde{\mathcal{I}}_k} L^{\text{local}}(x^{(n)}),$ which  controls the worst-case change in confidence for all samples that remain robustly inside bin $k$.

\noindent \textbf{Certified bin-wise calibration under perturbations.} For samples in $\widetilde{\mathcal{I}}_k$, let $A_k(\delta)$ and
$P_k(\delta)$ denote the accuracy and mean confidence under perturbation:
\begin{align}
  A_k(\delta)
  &= \frac{1}{|\widetilde{\mathcal{I}}_k|}
     \sum_{n\in\widetilde{\mathcal{I}}_k}
     \mathds{1}\{y^{(n)} = \hat{y}(x^{(n)}+\delta)\}, \nonumber \\
  P_k(\delta)
  &= \frac{1}{|\widetilde{\mathcal{I}}_k|}
     \sum_{n\in\widetilde{\mathcal{I}}_k}
     p_{\hat{y}(x^{(n)}+\delta)}(x^{(n)}+\delta). \nonumber
\end{align}
We can now decompose the bin-wise miscalibration by adding and subtracting the empirical $\hat{A}_k$ and $\hat{P}_k$ and applying the triangle inequality twice:
\begin{equation}
\label{eq:triangle_ineq}
  |A_k(\delta) - P_k(\delta)|
  \le |A_k(\delta) - \hat{A}_k|
     + |\hat{A}_k - \hat{P}_k|
     + |\hat{P}_k - P_k(\delta)|.    
\end{equation}
This decomposition isolates the three sources of uncertainty: statistical, empirical, and noise-based, and allows us to bound each one separately.
\begin{itemize}
  \item \textbf{(i) Statistical deviation:} 
  $|A_k(\delta) - \hat{A}_k|$ measures how much the \emph{true} accuracy under
  perturbations may differ from the empirical accuracy computed on the clean
  dataset. Since $A_k(\delta)$ depends on the model predictions for all
  $\delta$ in the perturbation ball, it cannot be computed directly and must
  be bounded using concentration inequalities, yielding a statistical term
  $\epsilon_k^{\mathrm{stat}}$. In particular, we resort to Hoeffding’s inequality \cite{hoeffding1963} to compute $\epsilon_k^{\mathrm{stat}}$, which results in $\epsilon_k^{\mathrm{stat}}=\sqrt{\frac{1}{2m_k}\ln (\frac{2}{\alpha})},$ where $m_k$ is the number of samples per bin, i.e., $|\tilde{\mathcal{I}_k}|$, and $
  \alpha$ the probability of making an error, set to 0.05. 

  \item \textbf{(ii) Empirical miscalibration:}
  $|\hat{A}_k - \hat{P}_k|$ is exactly the empirical bin-wise calibration
  error computed on clean calibration data. This is the standard calibration quantity used in ECE and requires no additional approximation.

  \item \textbf{(iii) Lipschitz deviation:}
  $|\hat{P}_k - P_k(\delta)|$ captures how much the mean confidence in the bin
  may change under perturbations. This term is controlled by the Lipschitz
  constant of the confidence function, leading to the bound $|\hat{P}_k - P_k(\delta)| \le L_k R$ (Eq. \ref{eq:lip_bound} and Eq. \ref{eq:gamma}).
\end{itemize}

\noindent \textbf{Proposed upper bound to certify confidence calibration.} Combining these terms yields the proposed
certified bin-wise confidence calibration guarantee. Specifically, for every admissible
perturbation $\delta$ with $\|\delta\|_2 \le R$, we obtain
\begin{equation}
  |A_k(\delta) - P_k(\delta)|
  \le
  \widehat{\mathrm{Cal}}_k
  + \epsilon_k^{\text{stat}}
  + L_k R,
  \qquad
  \forall\,\|\delta\|_2 \le R.
  \label{eq:proposed_bound}
\end{equation}
This bound certifies that, for all samples whose confidence remains within the robust sub-bin $\widetilde{B}_k$, the worst-case bin-wise miscalibration
cannot exceed the sum of the empirical calibration error, a statistical uncertainty term, and a Lipschitz-controlled perturbation term.

\section{Experiments}
\label{sec:experiments}

\subsection{Experiment setup}
\noindent \textbf{Models.} We use Vision Transformers (ViTs) \cite{dosovitskiyimage} as our main backbone, given their strong performance and enhanced calibration. Their structured size variants (Ti, S, B) also let us study how model capacity affects certified calibration.

\noindent \textbf{Datasets.} We use MedMNIST~\cite{medmnistv2}, a standardized medical image classification benchmark to evaluate \ours{}. In particular, we use the 2-dimensional datasets and exclude ChestMNIST because it is a multi-label task, leaving us with 11 total datasets. We use the provided training, validation, and test splits, where we use the validation split for certification and report the results on the test set. We note that \cite{bilic2023liver} reports that the train and test, which are used to build OrganA (OA), OrganC (OC) and OrganS (OS) contain some differences. Thus, we split the datasets in \textit{ID}: Blood (Bl), Breast (Br), Derma (D), OCT (O), Pathology (P), Pneumonia (Pn), Retina (R) and Tissue (T), and \textit{OOD}: OA, OC and OS.

\noindent \textbf{Baselines.} We compare \ours{} against two baselines: \textit{\Ltwo,} where the model is calibrated on $\ell_2$-perturbed data, and the bin-wise calibration error computed is used as the upper bound, regardless of the attack used. Similarly, we define ``attack-specific calibration'' (ASC), where the empirical calibration upper bound is computed with data perturbed with the specific adversarial attack used. \textbf{Adversarial attacks.} To evaluate the robustness of \ours{} we resort to well-known adversarial attacks: FGSM~\cite{goodfellow2014explaining}, BIM \cite{kurakin2018adversarial}, PGD~\cite{madry2018towards} and RFGSM \cite{tramèr2020ensemble}. In addition, we report results with bounded Gaussian noise, denoted as L2. For all perturbations, we use an $\ell_2$\text{-}certified radius $R=0.1$.

\noindent \textbf{Implementation details.} Since several bins sometimes contain too few samples to yield stable estimates, we approximate bin-specific $\epsilon_k$ by a global $\epsilon$.

\noindent \textbf{Evaluation metrics.} $\mathrm{Coverage}=\frac{1}{K}\sum_k\mathds{1}\big\{\mathrm{Cal}_{k,\text{test}} \leq \mathrm{UB}_k \big\}$ represents the fraction of bins for which the bound is respected, where $\mathrm{UB}_k$ is the upper bound for bin $k$ computed for each method. We define by $\mathrm{Tightness}=\frac{1}{K}\sum_k\vert \mathrm{Cal}_{k,\text{test}} - \mathrm{UB}_k  \vert$, the mean distance between the bound and the test calibration across bins.

\begin{table}[t!]
    \centering
    \scriptsize
    \caption{\textbf{Coverage and tightness.} Results across three ViT sizes and five attack types. $\boldsymbol{\mathcal{FT}}$ indicates if the methods requires specific noise fine-tuning.} 
    \label{tab:main}
    \resizebox{\textwidth}{!}{
    \begin{tabular}{
    >{\centering\arraybackslash}m{0.6cm}     
    >{\centering\arraybackslash}m{1.2cm}   
    >{\raggedright\arraybackslash}m{1.2cm}   
    >{\centering\arraybackslash}m{0.4cm}     
    >{\centering\arraybackslash}m{0.9cm}     
    >{\centering\arraybackslash}m{0.9cm}     
    >{\centering\arraybackslash}m{0.9cm}     
    >{\centering\arraybackslash}m{0.9cm}     
    >{\centering\arraybackslash}m{0.9cm}     
    >{\centering\arraybackslash}m{0.9cm}     
    >{\centering\arraybackslash}m{0.9cm}     
    >{\centering\arraybackslash}m{0.9cm}     
    >{\centering\arraybackslash}m{0.9cm}     
    >{\centering\arraybackslash}m{0.9cm}     
}
    \toprule
        & & & & \multicolumn{5}{c}{\textbf{Coverage}} & \multicolumn{5}{c}{\textbf{Tightness}} \\
        \cmidrule(lr){5-9} \cmidrule(lr){10-14}
        & & \textbf{Method} & \textbf{$\boldsymbol{\mathcal{FT}}$} & L2 & FGSM & BIM & RFGSM & PGD & L2 & FGSM & BIM & RFGSM & PGD \\
    \midrule
        \multirow{10}{*}{ID} & \multirow{3}{*}{ViT-Ti} & \Ltwo{} & \ding{52} & 0.36 & 0.34 & 0.36 & 0.34 & 0.34 & 0.11 & 0.30 & 0.18 & 0.32 & 0.32 \\
        & & ASC & \ding{52} & 0.36 & 0.41 & 0.41 & 0.44 & 0.44 & 0.11 & 0.21 & 0.17 & 0.22 & 0.22 \\
        & & \our \ours{} & \our \ding{56} & \our 0.99 & \our0.91 & \our0.98 & \our0.91 & \our0.91 & \our0.33 & \our0.34 & \our0.45 & \our0.32 & \our0.32 \\
        \cmidrule{2-14} & \multirow{3}{*}{ViT-S} & \Ltwo{} & \ding{52} & 0.38 & 0.36 & 0.38 & 0.36 & 0.36 & 0.12 & 0.26 & 0.19 & 0.27 & 0.27 \\
        & & ASC & \ding{52} & 0.38 & 0.44 & 0.38 & 0.46 & 0.48 & 0.12 & 0.16 & 0.17 & 0.18 & 0.18 \\
        & & \our \ours{} & \our \ding{56} & \our 0.98 & \our 0.90 & \our 0.98 & \our 0.90 & \our 0.90 & \our 0.39 & \our 0.44 & \our 0.50 & \our 0.43 & \our 0.43 \\
        \cmidrule{2-14} & \multirow{3}{*}{ViT-B} & \Ltwo{} & \ding{52} & 0.36 & 0.36 & 0.38 & 0.36 & 0.36 & 0.12 & 0.34 & 0.19 & 0.36 & 0.36 \\
        & & ASC & \ding{52} & 0.36 & 0.45 & 0.39 & 0.45 & 0.45 & 0.12 & 0.23 & 0.16 & 0.23 & 0.23 \\
        & & \our \ours{} & \our \ding{56} & \our 0.98 & \our 0.91 & \our 0.98 & \our 0.91 & \our 0.91 & \our 0.38 & \our 0.42 & \our 0.53 & \our 0.40 & \our 0.40 \\
    \midrule
        \multirow{10}{*}{OOD} & \multirow{3}{*}{ViT-Ti} & \Ltwo{} & \ding{52} & 0.20 & 0.20 & 0.20 & 0.20 & 0.20 & 0.18 & 0.43 & 0.24 & 0.45 & 0.45 \\
        & & ASC & \ding{52} & 0.20 & 0.37 & 0.20 & 0.37 & 0.37 & 0.18 & 0.17 & 0.23 & 0.20 & 0.20 \\
        & & \our \ours{} & \our \ding{56} & \our 0.60 & \our0.33 & \our0.60 & \our0.27 & \our0.27 & \our0.21 & \our0.26 & \our0.24 & \our0.29 & \our0.29 \\
        \cmidrule{2-14} & \multirow{3}{*}{ViT-S} & \Ltwo{} & \ding{52} & 0.20 & 0.20 & 0.20 & 0.20 & 0.20 & 0.15 & 0.44 & 0.23 & 0.47 & 0.47 \\
        & & ASC & \ding{52} & 0.20 & 0.40 & 0.20 & 0.37 & 0.37 & 0.15 & 0.21 & 0.23 & 0.23 & 0.23 \\
        & & \our \ours{} & \our \ding{56} & \our 0.93 & \our 0.40 & \our 0.80 & \our 0.40 & \our 0.40 & \our 0.30 & \our 0.23 & \our 0.28 & \our 0.24 & \our 0.24 \\
        \cmidrule{2-14} & \multirow{3}{*}{ViT-B} & \Ltwo{} & \ding{52} & 0.20 & 0.20 & 0.20 & 0.20 & 0.20 & 0.15 & 0.41 & 0.19 & 0.44 & 0.44 \\
        & & ASC & \ding{52} & 0.20 & 0.43 & 0.20 & 0.43 & 0.43 & 0.15 & 0.20 & 0.17 & 0.24 & 0.21 \\
        & & \our \ours{} & \our \ding{56} & \our 0.77 & \our 0.73 & \our 0.77 & \our 0.73 & \our 0.73 & \our 0.98 & \our 0.94 & \our 1.06 & \our 0.94 & \our 0.94 \\
    \bottomrule
    \end{tabular}}
\end{table}

\subsection{Results}
\noindent \textbf{\textit{Main results} (Table \ref{tab:main}).} \textit{\underline{ID scenario:}} Across all architectures and adversarial attacks, our method yields the highest coverage, between \textbf{0.90-0.99}, representing a \textbf{2-3$\boldsymbol{\times}$ increase} over the baselines, which hover around 0-30-0.44. Importantly, this improvement does not typically come at the cost of overly conservative or loose bounds. Furthermore, while ASC sometimes yields tighter estimates, it fails to provide useful coverage, and requires retraining for each noise level and attack, making it impractical. In contrast, \textbf{\ours{} certifies substantially more samples, while maintaining competitive tightness, all without requiring costly per‑noise or per‑attack retraining.} \textit{\underline{OOD scenario:}} As expected, performance drops under distribution shift. Nonetheless, \ours{} still provides the strongest guarantees, maintaining valid certifications in most cases, e.g., \textbf{mean coverage often $\geq$ 0.6, with a highest value of 0.93,} whereas the highest coverage of baselines is 0.43. Interestingly, \ours{} bounds tend to appear tighter (or at least comparable to the baselines) in this setting, largely due to the mismatch between the validation data (used to estimate empirical miscalibration) and the OOD test distribution. While this effect should not be interpreted as a genuine advantage, since it stems from a violation of the i.i.d. assumption underlying our certificate, \ours{} still achieves the overall highest certified coverage among the evaluated approaches. Finally, \textit{coverage} and \textit{tightness} must be interpreted jointly: a certificate that applies to only a small fraction of samples can appear tight, but is of limited practical value, whereas \ours{} 
provides meaningful, data-dependent bounds for most of the inputs under a wide spectrum of adversarial perturbations. 

\begin{figure}[t!]
    \centering
    \includegraphics[width=\linewidth]{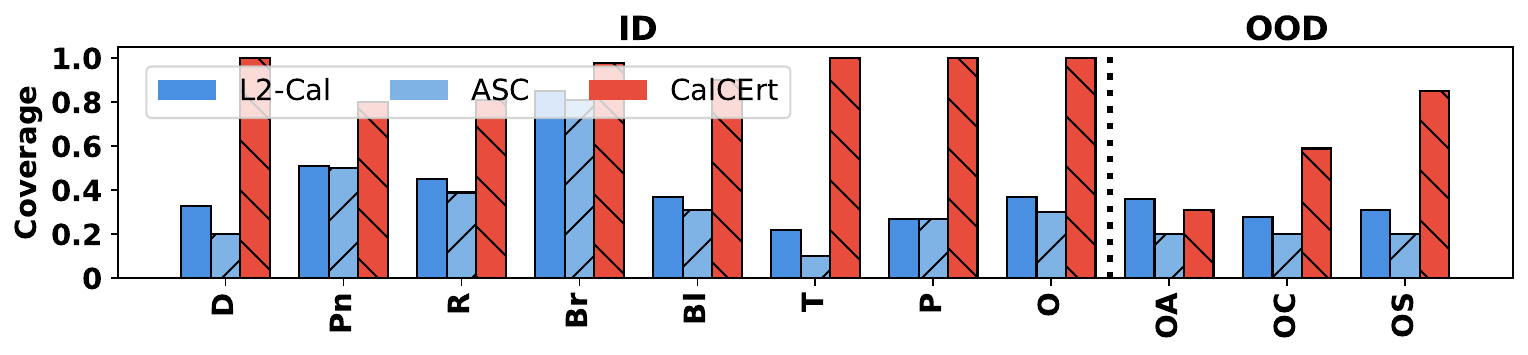}
    \caption{\textbf{Dataset-wise coverage.} Coverage for \Ltwo{}, ASC, and \ours{} across 11 MedMNIST datasets. Results are averaged across three ViT sizes.}
    \label{fig:coverage}
\end{figure}

\noindent \textbf{\textit{Per-dataset results} (Figure~\ref{fig:coverage}).} We observe that \ours{} consistently yields strong coverage gains across all ID datasets, \textbf{achieving perfect coverage in five out of 8 datasets}, and clearly outperforming both baselines. Under domain shift, improvements become more modest and occasionally only competitive, as in the OA dataset, reflecting the expected degradation when validation and test distributions diverge. Still, even in the OD regime, \ours{} remains the only method that preserves meaningful coverage rather than collapsing entirely.

\begin{figure}[t]
    \centering
   \includegraphics[width=0.485\textwidth]{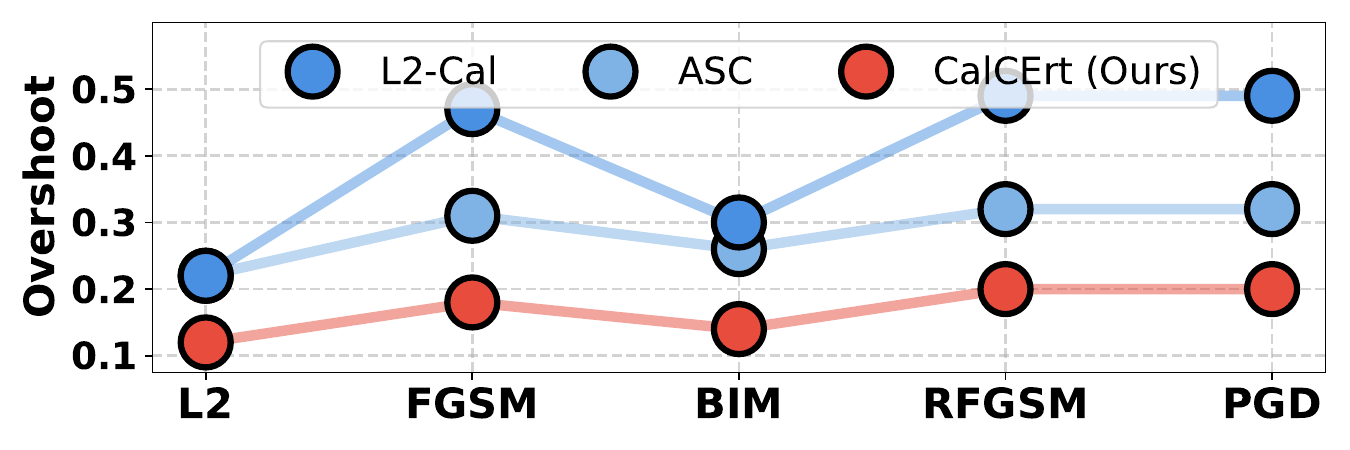}
    \caption{\textbf{Overshoot ($\downarrow$) across types of attacks:} mean across ViT backbones.}
       \label{fig:overshoot}
\end{figure}

\noindent \textbf{\textit{What happens when the bounds are violated?} (Figure~\ref{fig:overshoot}).} We define \textit{overshoot} as the mean positive gap between the true bin-wise miscalibration and the certified bound, evaluated over all bins where the bound is violated. In addition to its superior coverage, \textbf{\ours{} exhibits much smaller violation magnitudes whenever the bound is not satisfied.}

\noindent \textbf{\textit{Robustness across models} (Table~\ref{tab:RN}).} Consistent with our ViT results, \ours{} achieves \textbf{near‑full coverage} in the ID setting even when using a ResNet‑18 backbone. Under distributional shift, however, ResNet‑18 exhibits a more pronounced degradation. This is expected, as compared to ViTs, ResNets generally have lower representational capacity and weaker calibration, making their confidence estimates more sensitive to OOD drift. Moreover, because our certificate depends on bin‑wise confidence stability, architectures whose confidence distributions shift more under OOD conditions naturally yield fewer samples that remain within robust sub‑bins. Despite this drop, \ours{} still substantially outperforms the baselines in ID and remains competitive in OOD. 

\begin{table}[t!]
    \centering
    \scriptsize
    \caption{\textbf{Results with ResNet18 as backbone.}}
    \label{tab:RN}
    \resizebox{\textwidth}{!}{
    \begin{tabular}{
    >{\centering\arraybackslash}m{0.6cm}     
    >{\raggedright\arraybackslash}m{1.2cm}   
    >{\centering\arraybackslash}m{0.4cm}     
    >{\centering\arraybackslash}m{0.9cm}     
    >{\centering\arraybackslash}m{0.9cm}     
    >{\centering\arraybackslash}m{0.9cm}     
    >{\centering\arraybackslash}m{0.9cm}     
    >{\centering\arraybackslash}m{0.9cm}     
    >{\centering\arraybackslash}m{0.9cm}     
    >{\centering\arraybackslash}m{0.9cm}     
    >{\centering\arraybackslash}m{0.9cm}     
    >{\centering\arraybackslash}m{0.9cm}     
    >{\centering\arraybackslash}m{0.9cm}     
}
    \toprule
        & & & \multicolumn{5}{c}{\textbf{Coverage}} & \multicolumn{5}{c}{\textbf{Tightness}} \\
        \cmidrule(lr){4-8} \cmidrule(lr){9-13}
        & \textbf{Method} & \textbf{$\boldsymbol{\mathcal{FT}}$} & L2 & FGSM & BIM & RFGSM & PGD & L2 & FGSM & BIM & RFGSM & PGD \\
    \midrule
        \multirow{3}{*}{ID} & \Ltwo{} & \ding{52} & 0.35 & 0.34 & 0.33 & 0.34 & 0.34 & 0.12 & 0.32 & 0.25 & 0.36 & 0.36 \\
        & ASC & \ding{52} & 0.35 & 0.41 & 0.39 & 0.43 & 0.43 & 0.12 & 0.22 & 0.24 & 0.27 & 0.27 \\
        & \our \ours{} & \our \ding{56} & \our 0.99 & \our 0.95 & \our 0.95 & \our 0.95 & \our 0.95 & \our 0.44 & \our 0.53 & \our 0.59 & \our 0.49 & \our 0.49 \\
    \midrule
        \multirow{3}{*}{OOD} & \Ltwo{} & \ding{52} & 0.20 & 0.20 & 0.20 & 0.20 & 0.20 & 0.15 & 0.40 & 0.19 & 0.43 & 0.43 \\
        & ASC & \ding{52} & 0.20 & 0.27 & 0.23 & 0.27 & 0.27 & 0.15 & 0.23 & 0.19 & 0.25 & 0.26 \\
        & \our \ours{} & \our \ding{56} & \our 0.57 & \our 0.27 & \our 0.50 & \our 0.23 & \our 0.23 & \our 0.17 & \our 0.30 & \our 0.20 & \our 0.32 & \our 0.32\\
    \bottomrule
    \end{tabular}}
\end{table}


\section{Discussion and limitations}
This work should be viewed as an exploratory first step toward certifying confidence calibration under adversarial perturbations. Along the way, we encountered several conceptual and empirical limitations that highlight important challenges and opportunities for future work.
\textbf{First}, our bin-wise certificates depend on the empirical distribution of samples within each confidence bin. In several datasets, some bins contain very few samples, leading to loose statistical bounds and, consequently, large certified miscalibration values. Adaptive binning, where bin counts are equalized, could mitigate this imbalance, but we observed that it often produces \textit{near-overlapping} bins, which may complicate the construction of robust sub-bins, and violate the assumptions required for our certificates.
\textbf{Second,} \ours{} relies on local Lipschitz estimates of the confidence function. While this provides a principled way to control confidence variation under perturbations, it inherits well-known limitations of Lipschitz-based robustness methods. In particular, local Lipschitz constants can be large for some architectures, which increases the perturbation term in the certified bound and may render the certificate overly conservative or even vacuous. This reflects a broader challenge in Lipschitz-based certification: global bounds (e.g., products of spectral norms) are computationally efficient but notoriously loose \cite{leino2021globally,tsuzuku2018lipschitz}, whereas tighter local \cite{hashemi2021certifying} or semidefinite programming (SDP)-based \cite{fazlyab2019efficient} bounds are more accurate but scale poorly and are impractical for modern architectures. Our approach sits within this landscape and therefore shares these structural trade-offs. Furthermore, techniques such as spectral normalization, Jacobian penalties, or explicitly training low-Lipschitz networks \cite{anil2019sorting,pauli2021training} could reduce these constants and yield tighter certificates, but integrating such training-time strategies is incompatible with our post-hoc, model-agnostic setting.
\textbf{And third}, our certificates address only the robustness of confidence calibration and do not guarantee classification accuracy under adversarial perturbations. A model may remain well-calibrated while still misclassifying perturbed inputs, and conversely, accuracy robustness does not imply calibration robustness. Combining both within a unified framework is non-trivial, as the two objectives interact in complex ways and may require fundamentally different bounding strategies. Developing joint guarantees is therefore an important direction, but lies beyond the scope of this study. These limitations do not alter the main message of this work, where our goal is not to provide a definitive solution, but to explore how certification can be meaningfully extended to confidence calibration in safety‑critical settings.

\mypar{Acknowledgements.} This work has benefited from state financial aid, managed by the Agence Nationale de Recherche under the investment program integrated into France 2030, project reference ANR-21-RHUS-0003. This work was granted access to the HPC resources of IDRIS under the allocation 2025-AD011014802R1 made by GENCI. The author gratefully acknowledges support from ILLS during the internship at ETS Montreal.

\mypar{Disclosure of Interests.} The authors declare no competing interests.

\bibliographystyle{splncs04}
\bibliography{Paper-2687}

\end{document}

%% file: preamble.tex
\usepackage[dvipsnames]{xcolor}
\usepackage{color, colortbl}
\usepackage{amsmath}
\usepackage{amsfonts}
\usepackage{hyperref}
\usepackage{graphicx,verbatim}
\usepackage{booktabs}
\usepackage{multirow}
\usepackage{subcaption}
\usepackage{caption}
\usepackage{algorithm} 
\usepackage{algorithmic}
\usepackage{dsfont}
\usepackage{pifont}

\hypersetup{
    hidelinks
}

\newcommand{\mypar}[1]{\noindent\textbf{#1}\quad}

\definecolor{lightorange}{rgb}{0.98, 0.95, 0.92}
\newcommand{\our}{\cellcolor{lightorange}}

\definecolor{Better}{rgb}{0.18, 0.407, 0.266}
\definecolor{Worse}{rgb}{0.55, 0.22, 0.22}
\definecolor{Same}{rgb}{0.55, 0.55, 0.55}

\definecolor{myblue}{RGB}{38, 145, 199}

\newcommand{\ours}{\texttt{CalCErt}}

\newcommand{\Ltwo}{$L_2\textrm{--}\textsc{Cal}$}

\newcommand\blfootnote[1]{%
  \begingroup
  \renewcommand\thefootnote{}\footnote{#1}%
  \addtocounter{footnote}{-1}%
  \endgroup
}